\documentclass[a4paper]{esannV2}
\usepackage[dvips]{graphicx}
\usepackage[latin1]{inputenc}
\usepackage{amssymb,amsmath,array}

\usepackage{amsmath}
\usepackage{amsfonts}
\usepackage{amssymb}
\usepackage{algorithm}
\usepackage{algorithmic}
\usepackage{booktabs}
\usepackage{xspace}
\usepackage{xcolor}
\usepackage{caption}
\usepackage{subcaption}
\usepackage{fontawesome5}
\usepackage{pifont}
\usepackage{paralist}
\usepackage{enumitem}
\usepackage{hyperref}
\usepackage{cleveref}
\usepackage{nicefrac}
\usepackage{wrapfig}
\usepackage{bbm}
\usepackage{tikz}
\usetikzlibrary{
    shapes, 
    shapes.geometric, 
    shapes.arrows, 
    arrows, 
    patterns, 
    positioning, 
    calc,
}

\usepackage{pgfplots}
\usepgfplotslibrary{groupplots,colormaps}
\usetikzlibrary{pgfplots.colorbrewer}

\tikzset{every mark/.append style={mark size=1pt}}

\pgfplotsset{
    label style={font=\footnotesize},
    tick label style={font=\footnotesize}, 
    title style={font=\footnotesize, yshift=-1.2ex},
    ylabel shift=-1ex,
    xlabel shift=-0.8ex,
    axis lines*=left,
    grid=major,
    grid style={line width=.1pt, draw=gray!20, dashed},
    legend style={
        font = \scriptsize,
        fill = white!90, 
        draw opacity = 1,
        text opacity = 1,
        very thin,
        draw=none, 
    },
    compat=newest,
}

\pgfplotsset{
    barPlotStyle/.style={
        ybar,
        enlarge y limits = 0.05,
        enlarge x limits = 0.1,
        x tick label style={rotate=20, anchor=east},
	    cycle list/.style={
            colormap/Paired,
            cycle list/Paired,
        },
        every axis plot/.append style={fill},
        label style={font=\footnotesize},
        legend columns = -1,
        legend style={
            at = {(0.5, 1.15)}, 
            anchor = north,
        },
        height=3.5cm,
        width=0.98\textwidth,
        bar width=.25cm, 
    }
}

\pgfplotsset{
    linePlotStyle/.style={
        every axis plot/.append style={line width=1.8pt},
        legend style = {legend columns = -1},
        colormap/Paired,
        cycle list/Paired,
        cycle multiindex* list={
            mark list*\nextlist
            Paired\nextlist
            linestyles\nextlist
        },
        label style={font=\footnotesize},
        height=3.5cm,
    }
}

\hypersetup{
    pdfpagemode={UseOutlines},
    hypertexnames=false,
    colorlinks,
    linkcolor={blue},
    citecolor={blue},
    pdfstartview={FitH},
    pdftitle={End-to-End Neural Network Training for Hyperbox-Based Classification},
}

\definecolor{gcolor1}{HTML}{1b9e77}
\definecolor{gcolor2}{HTML}{d95f02}
\definecolor{gcolor3}{HTML}{7570b3}

\definecolor{darkgreen}{RGB}{0,160,0}
\definecolor{darkred}{RGB}{200,0,0}

\newcommand{\ie}{i.\nolinebreak[4]\hspace{0.01em}\nolinebreak[4]e.\@\xspace}
\newcommand{\eg}{e.\nolinebreak[4]\hspace{0.01em}\nolinebreak[4]g.\@\xspace}

\newcommand{\fscore}{$F_1$-score\@\xspace}
\newcommand{\sysname}{HyperNN\xspace}

\newcommand{\cmark}{\textcolor{darkgreen}{\ding{51}}}
\newcommand{\xmark}{\textcolor{darkred}{\ding{55}}}

\newcommand{\Reals}{{\mathbb R}}

\renewcommand{\vec}[1]{\mathbf{#1}}
\newcommand{\tdim}{d}
\newcommand{\tsize}{N}

\newcommand{\trainset}{T}
\newcommand{\clsize}{c}
\newcommand{\boxset}{\mathcal{B}}

\newcommand{\lossfunction}{\mathcal{L}}
\newcommand{\minboxpoints}{\boldsymbol{\theta}_m}
\newcommand{\lengthsbox}{\boldsymbol{\theta}_l}
\newcommand{\numboxes}{M}
\newcommand{\hypersmoothmax}{\phi}
\newcommand{\hypersigmoid}{\tau}

\newcommand{\smoothsigmoid}{\sigma_\hypersigmoid}
\newcommand{\smoothmaximum}{\mathcal{S}_\hypersmoothmax}

\newcommand{\traintime}{\mathcal{T}_{train}}
\newcommand{\predtime}{\mathcal{T}_{pred}}

\begin{document}
\title{End-to-End Neural Network Training for Hyperbox-Based Classification}

\author{Denis Mayr Lima Martins, Christian L{\"u}lf, Fabian Gieseke
%
%
\vspace{.3cm}\\
%
University of M{\"u}nster, ERCIS - Department of Information Systems\\ 
Leonardo Campus 3, 48149 - M{\"u}nster, Germany
%
}

\maketitle

\begin{abstract}
Hyperbox-based classification has been seen as a promising technique in which decisions on the data are represented as a series of orthogonal, multidimensional boxes (i.e., hyperboxes) that are often interpretable and human-readable. However, existing methods are no longer capable of efficiently handling the increasing volume of data many application domains face nowadays. We address this gap by proposing a novel, fully differentiable framework for hyperbox-based classification via neural networks. In contrast to previous work, our hyperbox models can be efficiently trained in an end-to-end fashion, which leads to significantly reduced training times and superior classification results.
\end{abstract}

\section{Introduction}\label{sec:intro}
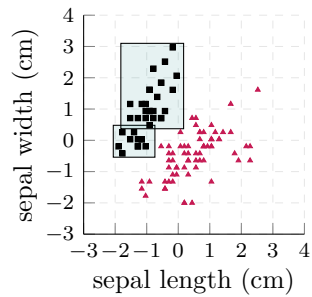
\begin{wrapfigure}{r}{0.47\textwidth}
    \vspace{-1.2em}
    \centering
    \resizebox{0.35\textwidth}{!}{\begin{tikzpicture}
\begin{axis}[
    width=4.5cm,
    height=4.5cm,
    xmin=-3, xmax=4,
    ymin=-3, ymax=4,
    xtick distance = 1,
    ytick distance = 1,
    xlabel={sepal length (cm)},
    ylabel={sepal width (cm)},
]
\addplot[ 
    only marks, 
    scatter,
    scatter src=explicit,
    scatter/classes={
        1={mark=square*, black}, 
        0={mark=triangle*, purple!90}
    },
]
table [x=x, y=y, meta=label]{%
x y label
-0.7884793081019493 2.291674296080614 1
1.2966104177676432 0.03227710276170067 0
0.0700870496090595 0.2582168220935923 0
-0.29786996083851647 -0.6455420552339742 0
1.0513057441359261 0.4841565414254829 0
0.4380440600566344 -2.0011803712253218 0
1.664567428215218 1.1619756994211579 0
-0.42052229765437443 -1.5493009325615397 0
0.19273938642491742 -0.41960233590208257 0
0.6833487336883514 0.2582168220935923 0
-0.052565287206799524 2.0657345767487234 1
-0.6658269712860914 1.3879154187530494 1
0.31539172324077536 -0.41960233590208257 0
-0.5431746344702324 0.7100962607573745 1
0.6833487336883514 0.03227710276170067 0
-1.4017409921812412 0.2582168220935923 1
-1.5243933289971001 0.03227710276170067 1
-1.1564363185495241 -1.3233612132296482 0
-0.9111316449178083 -1.3233612132296482 0
1.0513057441359261 0.03227710276170067 0
-1.1564363185495241 1.1619756994211579 1
1.2966104177676432 0.03227710276170067 0
-1.2790886553653833 -0.19366261657019096 1
-0.052565287206799524 -0.8714817745658648 0
0.19273938642491742 -0.8714817745658648 0
-1.769698002628816 0.2582168220935923 1
1.1739580809517842 0.2582168220935923 0
2.523133785926227 1.61385513808494 0
0.8060010705042093 -0.19366261657019096 0
2.155176775478651 -0.19366261657019096 0
0.5606963968724923 -0.6455420552339742 0
-1.5243933289971001 0.7100962607573745 1
-0.17521762402265745 -1.0974214938977565 0
1.0513057441359261 -1.3233612132296482 0
-0.17521762402265745 -0.6455420552339742 0
-1.1564363185495241 0.03227710276170067 1
0.6833487336883514 -0.6455420552339742 0
1.419262754583501 0.2582168220935923 0
-0.17521762402265745 -0.41960233590208257 0
-1.0337839817336663 1.1619756994211579 1
-0.5431746344702324 1.8397948574168317 1
-0.9111316449178083 1.61385513808494 1
-0.5431746344702324 -0.19366261657019096 0
1.9098721018469351 -0.6455420552339742 0
0.5606963968724923 -0.41960233590208257 0
0.9286534073200672 -0.19366261657019096 0
-1.0337839817336663 0.7100962607573745 1
-0.9111316449178083 1.61385513808494 1
-0.9111316449178083 0.9360359800892661 1
0.8060010705042093 -0.19366261657019096 0
-0.052565287206799524 -0.8714817745658648 0
0.31539172324077536 -0.6455420552339742 0
-0.42052229765437443 -1.5493009325615397 0
0.8060010705042093 0.2582168220935923 0
0.8060010705042093 -0.19366261657019096 0
0.0700870496090595 -0.19366261657019096 0
0.31539172324077536 -1.0974214938977565 0
0.6833487336883514 -0.41960233590208257 0
-0.7884793081019493 0.7100962607573745 1
-1.2790886553653833 0.7100962607573745 1
-0.29786996083851647 -0.8714817745658648 0
-0.9111316449178083 0.4841565414254829 1
-1.5243933289971001 1.1619756994211579 1
0.31539172324077536 -0.19366261657019096 0
-1.0337839817336663 -1.7752406518934314 0
-0.17521762402265745 1.61385513808494 1
-0.7884793081019493 0.9360359800892661 1
-0.42052229765437443 0.9360359800892661 1
-1.1564363185495241 -1.5493009325615397 0
-0.5431746344702324 0.7100962607573745 1
1.0513057441359261 0.03227710276170067 0
-1.2790886553653833 -0.19366261657019096 1
-0.42052229765437443 -1.3233612132296482 0
0.19273938642491742 -2.0011803712253218 0
-1.2790886553653833 0.03227710276170067 1
1.0513057441359261 0.03227710276170067 0
-0.052565287206799524 -1.0974214938977565 0
0.4380440600566344 0.7100962607573745 0
-1.0337839817336663 -0.19366261657019096 1
0.19273938642491742 -0.19366261657019096 0
-1.892350339444675 -0.19366261657019096 1
-0.29786996083851647 -1.3233612132296482 0
1.0513057441359261 -0.19366261657019096 0
2.27782911229451 -0.19366261657019096 0
0.5606963968724923 0.7100962607573745 0
-0.42052229765437443 2.5176140154125064 1
0.19273938642491742 -2.0011803712253218 0
2.27782911229451 -0.6455420552339742 0
1.787219765031076 -0.41960233590208257 0
-0.29786996083851647 -0.19366261657019096 0
0.8060010705042093 -0.6455420552339742 0
0.5606963968724923 0.4841565414254829 0
-0.5431746344702324 1.8397948574168317 1
0.6833487336883514 0.2582168220935923 0
-1.0337839817336663 0.9360359800892661 1
-0.9111316449178083 0.9360359800892661 1
0.31539172324077536 -0.6455420552339742 0
-1.769698002628816 -0.41960233590208257 1
1.0513057441359261 -0.19366261657019096 0
0.5606963968724923 -0.8714817745658648 0
-0.17521762402265745 -1.3233612132296482 0
-0.29786996083851647 -0.41960233590208257 0
-0.17521762402265745 2.96949345407629 1
1.664567428215218 0.2582168220935923 0
-1.1564363185495241 0.03227710276170067 1
};
\draw [solid, draw=black, fill=teal, fill opacity=0.1] (-2.06, -0.54) rectangle (-0.74, 0.48);
\draw [solid, draw=black, fill=teal, fill opacity=0.1] (-1.82,  0.37) rectangle (0.17,  3.1);
\end{axis}
\end{tikzpicture}}
    \vspace{-0.5em}
    \caption{Hyperbox-based classification for the Iris data set. Only a user-defined target class (black squares) is covered by two axes-aligned boxes.}
    \label{fig:example}
    \vspace{-1.2em}
\end{wrapfigure}

Hyperbox-based classification has been widely studied in the context of machine learning and data mining~\cite{HastieTF2009,Arzamasov2021reds,Khuat2021hyperbox}. The goal of the corresponding approaches is to identify/produce a set of hyperboxes (\ie, multidimensional rectangles) that collectively cover the data of interest (\eg, data points belonging to a class of interest in the context of classification scenarios)~\cite{Friedman1999prim}, as shown in \Cref{fig:example}. 
Using hyperboxes to represent regions of interest in the data has various advantages. One of them is that the resulting models can be interpreted more easily. For instance,
identifying such hyperboxes allows selecting representative data points or to provide user-friendly predicates/decision rules to describe objects belonging to a specific class. While there is no binary tree associated with such decisions, like it is the case for decision trees, the ``individual rules are often simpler''~\cite{HastieTF2009}.
Another advantage of simple predicates is the fact that they can give rise to orthogonal range queries in low-dimensional sub-spaces, which can efficiently be supported via indexing structures in the context of modern database management systems~\cite{FriedmannRana2021box}. 
These characteristics make hyperbox-based models promising alternatives to classic, opaque models (\eg, deep neural networks) for data-intense tasks in medicine, healthcare, pharmaceutical, and cybersecurity domains~\cite{Khuat2021hyperbox}.

Under existing approaches, 
\emph{patient rule induction method}~(PRIM)~\cite{HastieTF2009} and \emph{fuzzy min-max neural networks}~(FMMs)~\cite{Simpson1992FMM} have been the \textit{de facto} for hyperbox-based classification. 
These approaches are, however, not yet capable to cope with the increasing amounts of data many domains are confronted with.  
Also, one generally has little to no control 
over the number, size, and dimensionality of the induced hyperboxes.
In particular, current hyperbox-based neural networks~\cite{Khuat2021hyperbox} rely on non-differentiable modules, which prevents both end-to-end training via gradient-based optimization and the use of modern optimizers (see \Cref{tab:sota}). 

\begin{table}[t]
    \centering
    \caption{Comparison of hyperbox-based classification methods.}
    \resizebox{\textwidth}{!}{
        \begin{tabular}{l l c c c c}
            \toprule
            Approach & Training & Large $\tdim$ & Large $\tsize$ & End-to-end & Mult. hyperboxes\\
            \midrule
            PRIM~\cite{Friedman1999prim}             & Hill climbing        & \xmark & \xmark & \xmark & \cmark\\
            FMM~\cite{Simpson1992FMM} & Fuzzy membership     & \xmark & \xmark & \xmark & \cmark\\
            \textbf{\sysname (Ours)}                 & Gradient-based & \cmark & \cmark & \cmark & \cmark\\
            \bottomrule
        \end{tabular}
    }
    \label{tab:sota}
\end{table}

In this work, we introduce \sysname, a novel neural network for hyperbox-based classification method that can be trained in an end-to-end training fashion. We demonstrate via our experimental analysis that \sysname achieves a competitive if not superior classification performance compared to other state-of-the-art approaches, while reducing both training and inference times. 
Hence, to the best of our knowledge, this is the first work to propose a fully differentiable, end-to-end approach for hyperbox-based classification, which can be easily adapted via the use of appropriate loss functions and regularizers, and readily combined to modern deep neural networks (\eg, ResNets~\cite{He_2016_CVPR}) for enhanced classification. 

\section{Problem Formulation}\label{sec:background}


Given a $\tdim$-dimensional space, a hyperbox $B=B_{\minboxpoints,\lengthsbox} = \{\vec{x} \in \Reals^\tdim \mid \minboxpoints \leq \vec{x} \leq \minboxpoints + \lengthsbox\} \subset \Reals^\tdim$ can be characterized via its minimal point $\minboxpoints \in \Reals^\tdim$ along with a vector $\vec{0} \leq \lengthsbox \in \Reals^\tdim$ containing the length spans.
For a point $\vec{x} \in \Reals^\tdim$, let $\mathbbm{1}_{B}(\vec{x}) = 1~\mathrm{if}~\vec{x} \in B~\mathrm{and}~\mathbbm{1}_{B}(\vec{x}) = 0$, otherwise. 
%
Accordingly, for the union $\boxset = \bigcup^{\numboxes}_{k=1}B_k$ of $\numboxes$ hyperboxes $B_1,\ldots,B_\numboxes$, we have
$\mathbbm{1}_{\boxset}(\vec{x}) = \max(\mathbbm{1}_{B_1}(\vec{x}), \ldots, \mathbbm{1}_{B_\numboxes}(\vec{x}))$.

We consider binary classification tasks with training sets of the form $\trainset=\{(\vec{x}_1,y_1), \ldots, (\vec{x}_\tsize, y_\tsize)\} \subset \Reals^\tdim \times \{0,1\}$, where each instance $i$ is represented by a feature vector $\vec{x}_i$ and an associated class label $y_i$. The goal of the learning process is to find a set $B_1,\ldots,B_\numboxes$ of $\numboxes$ hyperboxes such that the binary classification model $\mathbbm{1}_\boxset : \Reals^\tdim \rightarrow \{0,1\}$ induced by the union $\boxset$ of those boxes minimizes 
    $G(\boxset) = \nicefrac{1}{\tsize} \sum_{i=1}^\tsize \lossfunction(\mathbbm{1}_{\boxset}(\vec{x}_i), y_i)$,
where $\lossfunction: \Reals \times \Reals \rightarrow \Reals$ is a suitable loss function. 
Here, we use the binary cross entropy~(BCE), 
which leads to 
    $G_{BCE}(\boxset) = -\nicefrac{1}{\tsize} \sum_{i=1}^{\tsize} y_{i} \log(\mathbbm{1}_{\boxset}(\vec{x}_i)) + (1 - y_{i}) \log(1 - \mathbbm{1}_{\boxset}(\vec{x}_i))$
as objective.

For the sake of simplicity, this work focuses on binary classification tasks and numerical features. However, our approach can be readily adapted to target other data types such as image and text (with an additional feature extraction step), or alternative tasks such as multi-class classification (by modifying $\lossfunction$).

\section{Differentiable Hyperbox-Based Classification}\label{sec:contribution}

\begin{figure}[t]
    \centering
    \begin{subfigure}[b]{0.49\textwidth}
        \centering
        \tikzstyle{inputneuron}=[circle, 
    draw = black, 
	fill = white,
	minimum size = 1.8em,
]
\tikzstyle{boxneuron}=[rectangle,
    rounded corners,
    draw = black, 
	fill=teal!20,
	minimum size = 2.8em,
]

\tikzstyle{outputneuron}=[circle, 
    draw = black, 
	fill = black!20,
	minimum size = 1.8em,
]

\tikzstyle{operation}=[rectangle, 
    draw = black, 
	fill = black!20,
	minimum size = 2.8em,
]

\newcommand{\inputnum}{5} 
 
\newcommand{\hiddennum}{4}  
 
\newcommand{\outputnum}{1} 
 
\begin{tikzpicture}[scale=0.6, transform shape, font=\Large]

    \draw[rectangle, dashed, draw=black!80] (1., -0.5) rectangle (5.3, -5.5);

    \foreach \i in {1,...,\inputnum}
    {
        \node[inputneuron] (Input-\i) at (0,-\i) {};
    }
     
    \foreach \i in {1,...,\hiddennum}
    {
        \ifthenelse{\i<\hiddennum}{
            \node[boxneuron,
                yshift=(\hiddennum-\inputnum)*5 mm,
            ] (Hidden-\i) at (3.0,-\i+0.25) {$h_{B_\i}$};
        }{
            \node[boxneuron,
                yshift=(\hiddennum-\inputnum)*5 mm,
            ] (Hidden-\i) at (3.0,-\i-0.25) {$h_{B_\numboxes}$};
        }
    }
     
    \foreach \i in {1,...,\outputnum}
    {
        \node[operation,
            yshift=(\outputnum-\inputnum)*5 mm
        ] (Output-\i) at (4.5,-\i) {$\smoothmaximum$};
    }
     
    \foreach \i in {1,...,\inputnum}
    {
        \foreach \j in {1,...,\hiddennum}
        {
            \draw[->, shorten >=1pt] (Input-\i) -- (Hidden-\j);   
        }
    }
     
    \foreach \i in {1,...,\hiddennum}
    {
        \foreach \j in {1,...,\outputnum}
        {
            \draw[->, shorten >=1pt] (Hidden-\i) -- (Output-\j);
        }
    }
    
    \foreach \i in {1,...,\inputnum}
    {
        \ifthenelse{\i<\inputnum}{
            \draw[<-, shorten <=1pt] (Input-\i) -- ++(-0.8,0) node[left]{$x_{\i}$};
        }{
            \draw[<-, shorten <=1pt] (Input-\i) -- ++(-0.8,0) node[left]{$x_{\tdim}$};
        }
    } 
    
    \foreach \i in {1,...,\outputnum}
    {            
        \ifthenelse{\outputnum>1}{
            \draw[->, shorten <=1pt] (Output-\i) -- ++(1,0)
            node[right]{$\hat{y}_{\i}$};
        }{
            \draw[->, shorten <=1pt] (Output-\i) -- ++(1.2,0)
            node[right]{$h_{\boxset}(\vec{x})$};    
        }
    }
    
    \node () at (-0.6, -\inputnum+0.65){$\vdots$};
    \node () at (3.0, -\inputnum+1.1){$\vdots$};
    \node () at (3, 0){$h_{\boxset}$};
\end{tikzpicture}
        \caption{\sysname Network Structure}
        \label{fig:nnarchitecture}
    \end{subfigure}
    \hfill
    \begin{subfigure}[b]{0.5\textwidth}
        \centering
        \newcommand{\minop}{$min$}
\newcommand{\smoothmax}{$S_\alpha$}
\newcommand{\smoothsigm}{$\sigma_\tau$}
\newcommand{\minparams}{$\boldsymbol{\theta}^{k}_{m}$}
\newcommand{\lenparams}{$\boldsymbol{\theta}^{k}_{l}$}
\newcommand{\maxparams}{$\boldsymbol{\theta}^{k}_{u}$}

\tikzstyle{inputneuron}=[circle, 
    draw = black, 
	fill = white,
	minimum size = 1.5em,
]

\tikzstyle{tensor}=[circle, 
    draw = black, 
	fill = black!20,
	minimum size = 3em,
]

\tikzstyle{operation}=[rectangle, 
    draw = black, 
	fill = black!20,
	minimum size = 2em,
]

\tikzstyle{arrow} = [->, shorten >=1pt]

\tikzstyle{customlabel} = [color=black]

\newcommand{\inputnum}{5} 

\begin{tikzpicture}[scale=0.6, transform shape, font=\Large]
    \foreach \i in {1,...,\inputnum}
    {
        \node[inputneuron] (inp-\i) at (-1,-\i+1) {};
    }

    \foreach \i in {1,...,\inputnum}
    {
        \ifthenelse{\i<\inputnum}{
            \draw[<-, shorten <=1pt] (inp-\i) -- ++(-0.8,0) node[left]{$x_{\i}$};
        }{
            \draw[<-, shorten <=1pt] (inp-\i) -- ++(-0.8,0) node[left]{$x_{\tdim}$};
        }
    }
    \draw[rectangle, rounded corners, draw=black!80, fill=teal!20] (-0.55, 0.49) rectangle (4.8, -4.9);
    \node[operation] (lowerthanmax) at (0.3, 0) {$-$};
    \node[operation, below right=3em of lowerthanmax] (maxs) {$+$};
    \node[tensor, below left=2em and 0em of maxs] (mins) {\minparams};
    \node[tensor, right=1em of mins] (len) {\lenparams};
    \node[operation, below=10.3em of lowerthanmax] (higherthanmin) {$-$};
    \node[operation, right=2em of lowerthanmax] (minop1) {\minop};
    \node[operation, right=2em of higherthanmin] (minop2) {\minop};
    \node[operation, right=2em of minop1] (sigmoid1) {\smoothsigm};
    \node[operation, right=2em of minop2] (sigmoid2) {\smoothsigm};
    \node[operation, above right=4em and 0.1em of sigmoid2] (mult) {$\times$};
    \node[right=1.5em of mult] (out) {$h_{B_k}(\vec{x})$};

    \draw[arrow] (mins) -- (maxs);
    \draw[arrow] (len) -- (maxs);
    \draw[arrow] (mins) -- (higherthanmin);
    \draw[arrow] (maxs) -- (lowerthanmax);
    \draw[arrow] (lowerthanmax) -- (minop1);
    \draw[arrow] (higherthanmin) -- (minop2);
    \draw[arrow] (minop1) -- (sigmoid1);
    \draw[arrow] (minop2) -- (sigmoid2);
    \draw[arrow] (sigmoid1) -- (mult);
    \draw[arrow] (sigmoid2) -- (mult);
    \draw[arrow] (mult.east) -- (out.west);

    \foreach \i in {1,...,\inputnum}
    {
        \draw[->, shorten >=1pt] (inp-\i) -- (higherthanmin);   
        \draw[->, shorten >=1pt] (inp-\i) -- (lowerthanmax);   
    }

    \node () at (-1.6, -4+0.65){$\vdots$};
    \node () at (2, 1){$h_{B_k}$};
\end{tikzpicture}
        \caption{Hyperbox Neuron}
        \label{fig:blockdiagram}
    \end{subfigure}
    \caption{Architecture of \sysname.}
\end{figure}

The \sysname architecture in \Cref{fig:nnarchitecture} is similar to the one introduced by Simpson~\cite{Simpson1992FMM}, where each neuron in the hidden layer represents a hyperbox characterized by two trainable weight vectors (\ie, model parameters) $\minboxpoints \in \Reals^\tdim$ and $\lengthsbox \in \Reals^\tdim$.
Such hidden neurons are named \emph{hyperbox neurons} thereafter. The number of neurons in the hidden layer corresponds to the maximum number of hyperboxes to be induced, which is controlled by a hyperparameter~$\numboxes$.

In a nutshell, the hidden layer is responsible to check for individual hyperbox containment, \ie, each hyperbox neuron checks whether a data instance is covered by its associated hyperbox. The output layer, in turn, consists of a single neuron that checks whether a data instance is contained in \emph{at least one} of the hyperboxes. 
The sequence of operations performed by each hyperbox neuron is depicted in \Cref{fig:blockdiagram}. We detail these operations next.


Let $h_{\boxset}$ be a \sysname network including $\numboxes$ hyperbox neurons $h_{B_1},\ldots, h_{B_\numboxes}$, see again \Cref{fig:nnarchitecture}. In a first step, for each hyperbox neuron $h_{B_k}, 1 \leq k \leq \numboxes$, upper hyperbox bounds are computed as $\boldsymbol{\theta}^{k}_{u} = {\boldsymbol{\theta}^{k}_{m}} + {\boldsymbol{\theta}^{k}_{l}}$, where ${\boldsymbol{\theta}^{k}_{m}}$ and ${\boldsymbol{\theta}^{k}_{l}}$ are the two trainable weight vectors of neuron $h_{B_k}$. Generally, a hyperbox containment check $h_{B_k}(\vec{x})$ for a data instance $\vec{x} = [x_1, \ldots, x_\tdim]^\top$ could be performed using $h_{B_k}(\vec{x}) = \mathbbm{1}_{B_k}(\vec{x})$. 
However, such an indicator function formulation would lead to a gradient of zero during backpropagation, which, in turn, would render gradient-based optimization not applicable. Instead, we implement the containment check by computing $\delta^k_{u}(\vec{x}) = \boldsymbol{\theta}^{k}_{u} - \vec{x}$ and $\delta^{k}_{m}(\vec{x}) = \vec{x} - \boldsymbol{\theta}^{k}_{m}$.

Note that, for $\vec{x}$ to be covered by the hyperbox represented by neuron $h_{B_k}$, both $\delta^{k}_{m}(\vec{x})$ and $\delta^k_{u}(\vec{x})$ must be non-negative for all the $\tdim$ dimensions. As before, in order to obtain meaningful gradient information in the backpropagation phase, we cannot resort to element-wise step functions to check for this property (\ie, $S_j(z) = 1$ if $z \geq 0$, and $S_j(z) = 0$ otherwise, for $j = 1, \ldots, \tdim$). Instead, we resort to a differentiable surrogate applied to the minimum value (across all $\tdim$ dimensions) of both $\delta^{k}_{m}(\vec{x})$ and $\delta^{k}_{u}(\vec{x})$, respectively. More precisely, for $\delta^{k}_{m}(\vec{x})$, we implement this check via a generalized sigmoid function: 

\begin{equation*}
    \smoothsigmoid(min(\delta^k_{m}(\vec{x}))) = \frac{1}{1+\exp(\nicefrac{-min(\delta^k_{m}(\vec{x}))}{\hypersigmoid})},
\end{equation*}

\noindent where $\hypersigmoid$ is a temperature hyperparameter that controls the smoothness of the containment check. Small values of $\hypersigmoid$ lead to an approximation to the original indicator function $\mathbbm{1}_{B_k}(\vec{x})$, while still providing valuable gradient information.
Accordingly, we implement the upper bound check via $\smoothsigmoid(min(\delta^k_{u}(\vec{x})))$. 

Hence, each hyperbox neuron outputs a value between $[0, 1]$ that expresses the degree of containment of $\vec{x}$ within its associated hyperbox.

Likewise, the neural network output $h_{\boxset}(\vec{x})$ must indicate whether at least one of the hyperboxes represented by the hidden neurons contains the input data point~$\vec{x}$. This could be achieved by simply taking the maximum over all the outputs $h_{B_1}(\vec{x}), \ldots, h_{B_K}(\vec{x})$. 

However, using the maximum only yields gradient information for a single box. Instead, we resort to a smooth maximum function $\smoothmaximum$ to conduct this step, where values close to one denote containment of $\vec{x}$, and $\hypersmoothmax$ controls smoothness of $\smoothmaximum$, as follows: 

\begin{equation*}\label{eq:smoothmax}
    \smoothmaximum(h_{B_1}(\vec{x}),\ldots, h_{B_\numboxes}(\vec{x}))= \frac{\sum _{k=1}^{\numboxes}h_{B_k}(\vec{x})\exp({\nicefrac{h_{B_k}(\vec{x})}{\hypersmoothmax}})}{\sum _{k=1}^{\numboxes}\exp({\nicefrac{h_{B_k}(\vec{x})}{\hypersmoothmax}})}.
\end{equation*}

Overall, we obtain meaningful gradient information via the simple, yet crucial modifications described above, which allows training the networks in an end-to-end fashion.

Training $h_{\boxset}$ involves finding, for each neuron $h_{B_k}$, suitable assignments for the associated weight vectors ${\boldsymbol{\theta}^{k}_{m}}$ and ${\boldsymbol{\theta}^{k}_{l}}$, 
in order to minimize the loss function introduced in \Cref{sec:background}. 

\section{Experiments and Results}\label{sec:experiments}

\begin{wraptable}{r}{0.44\textwidth}
    \vspace{-2cm}
    \centering
    \caption{Data Sets.}
    \label{tab:datasets}
    \resizebox{5.3cm}{!}{
    \begin{tabular}{l r r r}
        \toprule
        Data set & $\tsize$ & $\tdim$ & $\clsize$ \\ 
        \midrule
        \texttt{iris}       & 150     & 4   & 3  \\
        \texttt{wine}       & 178     & 13  & 3  \\
        \texttt{cancer}     & 569     & 30  & 2  \\
        \texttt{blood}      & 748     & 5   & 2  \\
        \texttt{cars}       & 1,728   & 6   & 4  \\
        \texttt{satimage}   & 6,430   & 36  & 6  \\
        \texttt{letter}     & 20,000  & 16  & 26 \\
        \texttt{sensit}    & 98,528  & 100 & 3  \\
        \texttt{covtype}    & 581,012 & 54  & 7  \\
        \bottomrule
    \end{tabular}
    }
    \vspace{-1cm}
\end{wraptable}

We report an experimental design and analysis on several benchmark datasets, with focus on
\begin{inparaenum}[(1)]
\item effectiveness of our approach in comparison to widely-used baselines;
\item efficiency in terms of training and inference times;
\item sensitivity to the number of hyperboxes ($\numboxes$).
\end{inparaenum}

\subsection{Experimental Design}

We consider nine data sets included in the UCI Repository~(see \Cref{tab:datasets}, where $\clsize$ denotes the number of distinct classes). We employ a ``one-versus-all'' strategy to transform the original task into a binary classification. We use the ratio $70/30$ to split the data into training and test sets, and evaluate all methods in terms of \fscore, training time ($\traintime$), and inference time ($\predtime$).

For comparison, we use the PRIM implementation provided by David Hadka\footnote{\url{https://github.com/Project-Platypus/PRIM}}, and the recent FMM implementation by Thanh Tung Khuat\footnote{\url{https://github.com/UTS-AAi/comparative-gfmm}}, while \sysname is implemented in Python/PyTorch\footnote{\url{https://github.com/mlde-ms/hypernn}}. 
In all experiments, we conduct hyperparameter tuning using grid search. 
Best performing models are selected via averaged \fscore over 5-fold cross-validation. 
We set the training epochs to $10,000$, with early stopping of $200$ epochs when no further improvement is achieved on a holdout validation data set. For \sysname, we use the Adam optimizer. 
All experiments are conducted on an Ubuntu 18.04~server with 24~AMD EPYC 7402P cores, 192~GB RAM, and NVIDIA GeForce RTX 3090~GPU. In contrast to \sysname, both PRIM and FMM \emph{do not} make use of a GPU for fast computations.

\subsection{Results}

\Cref{fig:mean-results} reports results averaged over three runs using different random seeds. Note that we do not report FMM results on the larger data sets, since training time has not been concluded after a pre-defined time limit of ten hours. 
Both PRIM and FMM achieves high classification performance in terms of \fscore for all data sets. For large data sets such as \texttt{satimage} and \texttt{sensit}, however, these results are produced at a cost of high training times. In contrast, \sysname shows similar classification performance while keeping lower training times for almost all data sets. For \texttt{satimage} and \texttt{sensit}, \sysname achieves an \fscore close to PRIM in a fraction of the training time of the latter.

\begin{figure}[t]
    \centering
    \pgfplotstableread[row sep=\\, col sep=&]{
	dataset     & ours	    & PRIM      & FMM   \\
	iris        & 0.972		& 0.910     & 0.938 \\
	wine        & 0.951	    & 0.839     & 0.855 \\
	cars        & 0.860 	& 0.560     & 0.563 \\
    cancer      & 0.905     & 0.865     & 0.913 \\
	blood       & 0.698     & 0.636     & 0.591 \\
    satimage    & 0.852     & 0.794     & 0.855 \\
    letter      & 0.833     & 0.593     & 0.000 \\
    sensit      & 0.780     & 0.750     & 0.000 \\
    covtype     & 0.609     & 0.170     & 0.000 \\
}\fscoretable
\pgfplotstableread[row sep=\\, col sep=&]{
	dataset     & ours	    & PRIM      & FMM   \\
	iris        & 0.755		& 0.040     & 0.017 \\
	wine        & 1.351	    & 0.153     & 7.167 \\
	cars        & 1.283 	& 0.194     & 1.982 \\
    cancer      & 1.000     & 1.316     & 0.124 \\
	blood       & 0.718     & 0.137     & 3.149 \\
    satimage    & 0.971     & 8.738     & 7.086 \\
    letter      & 1.029     & 2.277     & 0.000 \\
    sensit      & 7.849     & 1616.06   & 0.000 \\
    covtype     & 4.542     & 26.21     & 0.000 \\
}\trtimetable

\begin{tikzpicture}[scale=0.94, transform shape, font=\Large]
\begin{groupplot}[
        barPlotStyle,
        group style={%
            group size=1 by 2,
            vertical sep=1cm,
        },
        symbolic x coords={
            iris, 
            wine, 
            cars, 
            cancer, 
            blood, 
            satimage, 
            letter, 
            sensit, 
            covtype
        }, 
        width=0.99\linewidth,
    ]   
	\nextgroupplot[
        ymin = 0,
        ymax = 1,
        ytick distance=0.25,
        xlabel = {}, 
        ylabel = {\fscore},
    ]
    	\addplot[
            draw=gcolor1, 
            fill=gcolor1!90, 
            pattern=crosshatch, pattern color=gcolor1,
        ] table[x=dataset, y=ours]{\fscoretable};
		\addplot[
            draw=gcolor2, 
            fill=gcolor2!90, 
            pattern=north east lines, pattern color=gcolor2,
        ] table[x=dataset, y=PRIM]{\fscoretable};
		\addplot[
            draw=gcolor3, 
            fill=gcolor3!80,
            pattern=north west lines, pattern color=gcolor3,
        ] table[x=dataset, y=FMM]{\fscoretable};
        \node[rotate=90, shift={(0.0cm, -0.35cm)}] at (axis cs:letter, 0.25) {\scriptsize N/A};
        \node[rotate=90, shift={(0.0cm, -0.35cm)}] at (axis cs:sensit, 0.25) {\scriptsize N/A};
        \node[rotate=90, shift={(0.0cm, -0.35cm)}] at (axis cs:covtype, 0.25) {\scriptsize N/A};
        
        \legend{\sysname, PRIM, FMM}

    \nextgroupplot[
        ymin = 0,
        ymax = 40,
        xlabel = {}, 
        ylabel = {$\traintime$ (sec.)},
        ytick distance = 10,
    ]

    \addplot[
        draw=gcolor1, 
        fill=gcolor1!90, 
        pattern=crosshatch, pattern color=gcolor1,
    ] table[x=dataset, y=ours]{\trtimetable};
    \addplot[
        draw=gcolor2, 
        fill=gcolor2!90, 
        pattern=north east lines, pattern color=gcolor2,
    ] table[x=dataset, y=PRIM]{\trtimetable};
    \node[
        pin={[pin distance=1cm, 
        pin edge={<-,>=stealth'},
        shift={(-1.0cm, -1.8cm)}
    ] \scriptsize 1616.06}] at (axis cs:sensit, 40) {};
    \addplot[
        draw=gcolor3, 
        fill=gcolor3!80,
        pattern=north west lines, pattern color=gcolor3,
    ] table[x=dataset, y=FMM]{\trtimetable};
    \node[rotate=90, shift={(0.0cm, -0.35cm)}] at (axis cs:letter, 10) {\scriptsize N/A};
    \node[rotate=90, shift={(0.0cm, -0.35cm)}] at (axis cs:sensit, 10) {\scriptsize N/A};
    \node[rotate=90, shift={(0.0cm, -0.35cm)}] at (axis cs:covtype, 10) {\scriptsize N/A};
	\end{groupplot}
\end{tikzpicture}
    \caption{Mean \fscore~(above) and $\traintime$~(below) obtained in our experiments.}
    \label{fig:mean-results}
\end{figure}

We also explore how sensitive \sysname is to changes in its main hyperparameters. \Cref{fig:effect-nboxes} shows the effect of $\numboxes$ in terms of \fscore, $\traintime$, and $\predtime$, where \sysname shows a stable scalability and generalization performance for an increasing $\numboxes$. 
For small datasets, such as \texttt{iris}, \texttt{wine}, and \texttt{cancer}, increasing $\numboxes$ brings almost no benefit in terms of \fscore. In contrast, for \texttt{letter}, \texttt{sensit}, and \texttt{covtype}, a high $\numboxes$ rapidly improves classification performance, at a cost of higher training and prediction times. 
However, for \texttt{blood}, increasing $\numboxes$ from $10$ to $20$ decreases \fscore due to overfitting. Such a degradation in classification performance could be alleviated by, \eg, an adaptive training procedure where $\numboxes$ is adapted (\ie, increased or decreased) if the validation loss deteriorates.

\begin{figure}[t]
    \centering
    \pgfplotstableread[row sep=\\, col sep=&]{
	nboxes &   f1score &    traintime   &  predtime\\
    2   &   0.8995628322237744  &   0.9057477712631226  &   0.000472\\
    5   &   0.9032291933425669  &   1.1050256888071697  &   0.000553\\
    10  &   0.9173221488852851  &   0.9546878735224406  &   0.000476\\
    20  &   0.9234313844504866  &   1.129006822903951   &   0.000446\\
    30  &   0.920749489018598   &   1.0789997577667236  &   0.000569\\
}\bcancerboxes

\pgfplotstableread[row sep=\\, col sep=&]{
	nboxes    &     traintime    &     predtime    &     f1score\\
    2    &     1.2744916280110676    &     0.00043736563788519963    &     0.9489600437876299\\
    5    &     1.2548084259033203    &     0.00045238600836859807    &     0.9564071043075493\\
    10    &     1.210390912161933    &     0.00046433342827690975    &     0.940974882928906\\
    20    &     1.265438715616862    &     0.0005172093709309896    &     0.964300073407415\\
    30    &     1.5209195348951552    &     0.0004980034298366971    &     0.9602132279412591\\
}\irisboxes

\pgfplotstableread[row sep=\\, col sep=&]{
	nboxes    &     traintime    &     predtime    &     f1score\\
    2    &     0.537096897761027    &     0.0004970232645670573    &     0.6744160054076808\\
    5    &     0.648353377978007    &     0.0005501508712768555    &     0.6855042295895041\\
    10    &     0.7185836633046468    &     0.0004715124766031901    &     0.6980783067544777\\
    20    &     0.7551075220108032    &     0.0004897912343343099    &     0.6604525718106181\\
    30    &     0.7954114278157552    &     0.00045752525329589844    &     0.6761972202889593\\
}\bloodboxes

\pgfplotstableread[row sep=\\, col sep=&]{
	nboxes    &     traintime    &     predtime    &     f1score\\
    2    &     1.358162932925754    &     0.0004811816745334201    &     0.9345896587832071\\
    5    &     1.4652027024163141    &     0.00046141942342122394    &     0.9517361527589695\\
    10    &     1.5432064798143175    &     0.0005369981129964193    &     0.9456911715520335\\
    20    &     1.4644013510810003    &     0.000438690185546875    &     0.9483700635016424\\
    30    &     1.5349846680959065    &     0.0005447599622938368    &     0.9519968424694607\\
}\wineboxes

\pgfplotstableread[row sep=\\, col sep=&]{
	nboxes    &     traintime    &     predtime    &     f1score\\
    2    &     1.0693191091219585    &     0.0005359053611755371    &     0.6778074227605609\\
    5    &     1.248545269171397    &     0.0005249778429667155    &     0.7899733710642994\\
    10    &     1.226841668287913    &     0.0004911621411641439    &     0.8280564706930972\\
    20    &     1.2516805132230122    &     0.0005335410435994467    &     0.860479219656291\\
    30    &     1.4978731671969097    &     0.00048287709554036457    &     0.87214476322949\\
}\carsboxes

\pgfplotstableread[row sep=\\, col sep=&]{
	nboxes    &     traintime    &     predtime    &     f1score\\
    2    &     0.8688213427861532    &     0.00127104918162028    &     0.8226866178679789\\
    5    &     0.948282970322503    &     0.001567840576171875    &     0.8383062638136881\\
    10    &     0.9717322985331217    &     0.0020694202846950954    &     0.8523515029914878\\
    20    &     1.055382980240716    &     0.0013794501622517903    &     0.8579521104520736\\
    30    &     1.2597881688012018    &     0.0009323623445298937    &     0.8551537978858492\\    
}\satimageboxes

\pgfplotstableread[row sep=\\, col sep=&]{
	nboxes    &     traintime    &     predtime    &     f1score\\
    2    &     0.893030964411222    &     0.0017968446780473758    &     0.7476980106876556\\
    5    &     1.029514422783485    &     0.0012955848987285907    &     0.8334080528880565\\
    10    &     1.4729535029484675    &     0.0013715028762817383    &     0.8706085050033221\\
    20    &     1.5786812060918562    &     0.0016180735367995042    &     0.8886187958174282\\
    30    &     2.076009126809927    &     0.0015413149809226012    &     0.8991300389454441\\
}\letterboxes

\pgfplotstableread[row sep=\\, col sep=&]{
	nboxes    &     traintime    &     predtime    &     f1score\\
    2    &     0.8291808234320747    &     0.003466659122043186    &     0.7446976851473839\\
    5    &     1.536630974875556    &     0.0034093326992458766    &     0.7636680183746195\\
    10    &     2.7672192785474987    &     0.0037460062238905164    &     0.7753777325968608\\
    20    &     5.26028839747111    &     0.004729721281263564    &     0.7795910754843045\\
    30    &     8.278811348809135    &     0.006810347239176433    &     0.7822656213608019\\ 
}\sensitboxes

\pgfplotstableread[row sep=\\, col sep=&]{
	nboxes    &     traintime    &     predtime    &     f1score\\
    2    &     2.0540366626921154    &     0.01204614412216913    &     0.5473545107854664\\
    5    &     4.545245159239996    &     0.013285818554106214    &     0.6096141761939364\\
    10    &     9.537449995676676    &     0.015612375168573289    &     0.6441086255156709\\
    20    &     18.661890711103165    &     0.01926210948399135    &     0.6600540204287467\\
    30    &     34.0348090557825    &     0.022969790867396762    &     0.6959807444436902\\    
}\covtypeboxes

\begin{tikzpicture}[scale=0.95, transform shape, font=\Large]
    \begin{groupplot}[
        linePlotStyle,
        group style={%
            group size=3 by 1,
            horizontal sep=1.5cm,
        },
        legend columns=5,
        legend style={
            at = {(2.0, 1.5)}, 
            anchor = north,
        },
        xmin = 0,
        xmax = 30,
        width=0.35\linewidth,
    ]   
    \nextgroupplot[
        ymin = 0.5,
        ymax = 1,
        ytick distance = 0.1,
        mark repeat=1,
        ylabel={\fscore}
    ]
    \addplot+[] table[x=nboxes, y=f1score]{\irisboxes};
    \addplot+[] table[x=nboxes, y=f1score]{\wineboxes};
    \addplot+[] table[x=nboxes, y=f1score]{\carsboxes};
    \addplot+[] table[x=nboxes, y=f1score]{\bcancerboxes};
    \addplot+[] table[x=nboxes, y=f1score]{\bloodboxes};
    \addplot+[] table[x=nboxes, y=f1score]{\satimageboxes};
    \addplot+[] table[x=nboxes, y=f1score]{\letterboxes};
    \addplot+[] table[x=nboxes, y=f1score]{\sensitboxes};
    \addplot+[] table[x=nboxes, y=f1score]{\covtypeboxes};
    \legend {
        \texttt{iris},
        \texttt{wine},
        \texttt{cars},
        \texttt{cancer},
        \texttt{blood},
        \texttt{satimage},
        \texttt{letter},
        \texttt{sensit},
        \texttt{covtype},
    }
    
    \nextgroupplot[
        ymin = 0,
        ymax = 40,
        ytick distance = 10,
        mark repeat=1,
        ylabel = {$\traintime$ (sec.)},
        xlabel = {Number of Hyperboxes (\numboxes)},
    ]
    \addplot+[] table[x=nboxes, y=traintime]{\irisboxes};
    \addplot+[] table[x=nboxes, y=traintime]{\wineboxes};
    \addplot+[] table[x=nboxes, y=traintime]{\carsboxes};
    \addplot+[] table[x=nboxes, y=traintime]{\bcancerboxes};
    \addplot+[] table[x=nboxes, y=traintime]{\bloodboxes};
    \addplot+[] table[x=nboxes, y=traintime]{\satimageboxes};
    \addplot+[] table[x=nboxes, y=traintime]{\letterboxes};
    \addplot+[] table[x=nboxes, y=traintime]{\sensitboxes};
    \addplot+[] table[x=nboxes, y=traintime]{\covtypeboxes};

    \nextgroupplot[
        ymin = 0,
        ymax = 0.03,
        ytick distance = 0.01,
        mark repeat=1,
        ylabel = {$\predtime$ (sec.)},
    ]
    \addplot+[] table[x=nboxes, y=predtime]{\irisboxes};
    \addplot+[] table[x=nboxes, y=predtime]{\wineboxes};
    \addplot+[] table[x=nboxes, y=predtime]{\carsboxes};
    \addplot+[] table[x=nboxes, y=predtime]{\bcancerboxes};
    \addplot+[] table[x=nboxes, y=predtime]{\bloodboxes};
    \addplot+[] table[x=nboxes, y=predtime]{\satimageboxes};
    \addplot+[] table[x=nboxes, y=predtime]{\letterboxes};
    \addplot+[] table[x=nboxes, y=predtime]{\sensitboxes};
    \addplot+[] table[x=nboxes, y=predtime]{\covtypeboxes};
\end{groupplot}
\end{tikzpicture}
    \caption{Effect of $\numboxes$ on \fscore~(left), $\traintime$~(center), and $\predtime$~(right).}
    \label{fig:effect-nboxes}
\end{figure}

\section{Conclusion}\label{sec:conclusion}

We propose \sysname, a fully differential approach for hyperbox-based classification. We provide an efficient, GPU-ready implementation that produced highly competitive models in terms of both classification and runtime performance, when compared to state-of-the-art techniques such as PRIM and FMM. 
As future work, we plan to apply \sysname to image data, in combination with other modern deep learning models (\eg, CNNs, ResNets), where both suitable features and hyperboxes must be learned jointly in an end-to-end fashion. 



\begin{footnotesize}

\bibliographystyle{unsrt}
\bibliography{references}

\end{footnotesize}


\end{document}